\documentclass[runningheads]{llncs}

\usepackage{eccv}

\usepackage{eccvabbrv}

\usepackage{graphicx}
\usepackage{booktabs}

\usepackage[accsupp]{axessibility}  

\usepackage[table]{xcolor}
\usepackage{amssymb}
\usepackage{pifont}
\usepackage{arydshln}
\newcommand{\cmark}{\ding{51}}%
\newcommand{\xmark}{\ding{55}}%
\newcommand*\samethanks[1][\value{footnote}]{\footnotemark[#1]}

\usepackage{multicol}
\usepackage{multirow}

\usepackage{colortbl}

\usepackage{subcaption}

\usepackage{array}
\newcolumntype{V}{>{\centering\arraybackslash}m{1.6em}} 

\usepackage{hyperref}

\usepackage{orcidlink}

\begin{document}


\title{What is the Right Embedding Space for Contrastive Learning in Referring Expression Counting?}



\titlerunning{What is the Right Embedding Space for Contrastive Learning in REC?}

\author{Kostas Triaridis\thanks{Equal contribution. Correspondence to \href{mailto:kostas@cs.stonybrook.edu}{kostas@cs.stonybrook.edu} }\inst{1}\orcidlink{0009-0009-5868-9596} 
\and
Panagiotis Kaliosis\samethanks[1]\inst{1}\orcidlink{0009-0001-7115-0431} 
\and
E-Ro Nguyen\inst{1}\orcidlink{0000-0002-8450-5190}
\and
Jingyi Xu\inst{1}\orcidlink{0009-0003-5597-5534
}
\and\\
Dimitris Samaras\inst{1}\orcidlink{0000-0002-1373-0294}
\and
Hieu Le\inst{2}\orcidlink{0000-0001-7855-2778}
}

\authorrunning{K.~Triaridis et al.}

\institute{Stony Brook University \and UNC-Charlotte}

\maketitle

\textbf{Project Page:} \texttt{\href{https://cvlab-stonybrook.github.io/C-REX/}{https://cvlab-stonybrook.github.io/C-REX/}}

\begin{abstract}
Referring Expression Counting (REC) requires distinguishing visually similar objects described by fine-grained text cues.
Existing methods tackle this via image-text contrastive learning of visual features which aims to distinguish visual features corresponding to positive and negative referring expressions. However, this strategy relies heavily on accurate image–text alignment and is constrained by the limited number of available negative expressions.
We argue that these limitations can be avoided by performing contrastive learning entirely in the visual embedding space. To this end, we introduce C-REX, a simple but effective supervised contrastive learning framework that learns fine-grained visual representations by contrasting visual tokens within the same image. By shifting the negative samples from incorrect Referring Expressions to visual tokens from the image, C-REX vastly increases the number of potential negatives, providing richer and more stable supervision. This design leads to stronger fine-grained visual discrimination and better generalization across complex open-world counting settings. C-REX serves as a plug-in module that can be seamlessly applied to any existing REC model without architectural changes. We apply it in three different REC architectures and achieve state-of-the-art results improving by up to 28\% in MAE and 24.5\% in RMSE. Moreover, we show that our framework is versatile and general, and can be applied to other counting tasks like class-agnostic counting, improving the performance of existing models. 
\end{abstract}

\begin{figure}
    \centering
    \includegraphics[width=\linewidth]{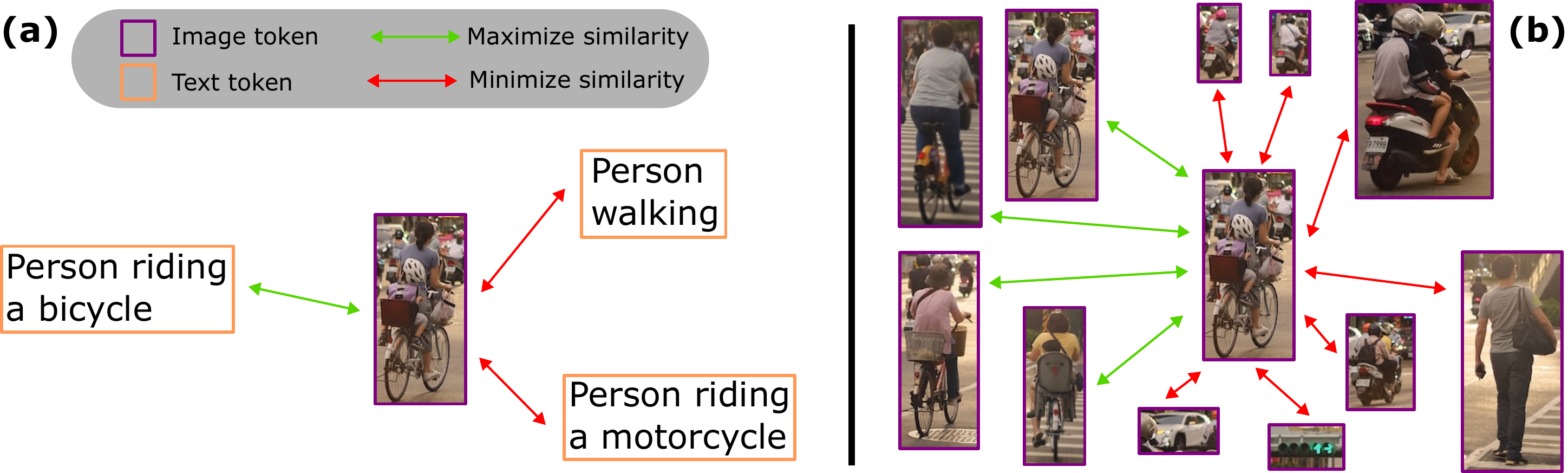}
    \caption{\textbf{(a)} 
    Standard REC models attempt to distinguish visually similar but contextually distinct objects through image-text contrastive learning. This approach can fall short when the image and text embedding spaces are poorly aligned.
    \textbf{(b)} We propose operating \textbf{solely} in the \textbf{image-space} for contrastive learning, directly optimizing relationships among visual embeddings. This provides stronger, more stable supervision, improving fine-grained contextual discrimination between visually similar objects.}
    \label{fig:teaser}
\end{figure}

\section{Introduction}
\label{sec:intro}

\begin{figure}[t]
    \centering
    \includegraphics[width=\textwidth]{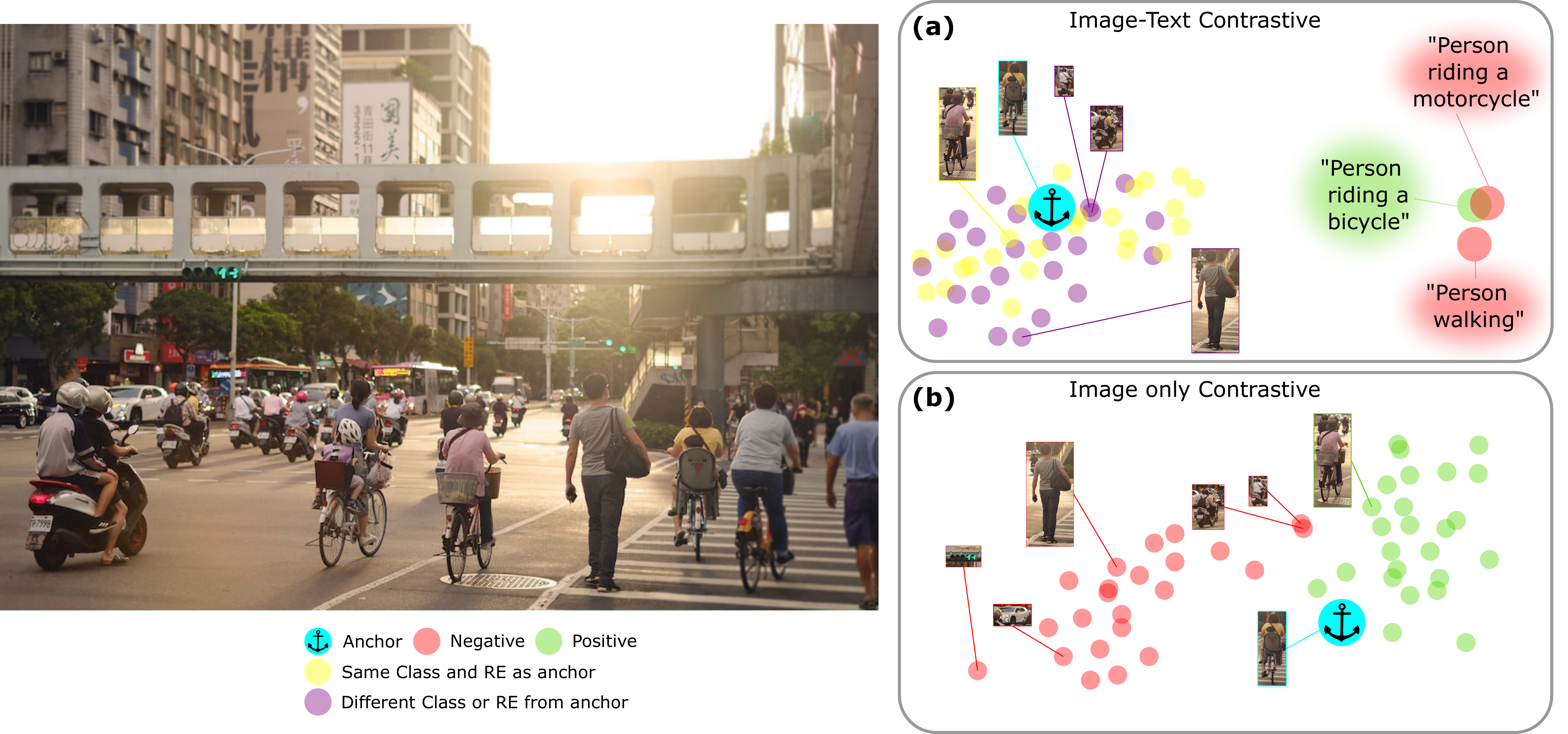}
    \caption{Existing REC methods \cite{groundingrec, cadgd} rely on image-text contrastive learning approaches that depend on text guidance, which can introduce misalignment and limit fine-grained separation between visually similar instances. \textbf{(a)} Each anchor visual token (in blue) is compared to the positive (green) and negative (red) prompt text embeddings. The image and text embedding spaces are not necessarily aligned which can lead to the visual tokens that correspond to those different prompts to be entagled, as they are not separated directly. \textbf{(b)} Instead, we propose performing \textbf{image-only} contrastive learning, directly optimizing relationships among object embeddings in the visual space. This yields more compact intra-class clusters and clearer inter-class separation, improving discrimination between visually and contextually distinct objects.}
    \label{fig:method}
\end{figure}

Referring Expression Counting (REC)~\cite{groundingrec} generalizes object counting from predefined classes to open-world settings, where the targets are specified by free-form referring expressions (REs) that describe both the object category and its fine-grained attributes. Instead of counting all instances of a predefined class, REC enables queries like “\textit{person walking},” or “\textit{person riding a motorcycle},” for the input image in Figures~\ref{fig:teaser} and~\ref{fig:method}. This formulation brings object counting closer to real-world scenarios, where users naturally express counting targets in language rather than selecting from a fixed set of labels.
However, this flexibility introduces a key challenge: developing robust, fine-grained visual representations that can precisely separate visually similar objects under diverse expressions.

Previous works~\cite{groundingrec, cadgd} address this challenge using image-text contrastive learning (CL).
The model is trained to distinguish visual objects associated with a positive prompt (e.g., ``person riding a bicycle'') from those paired with negative prompts (e.g., ''person riding a motorcycle'' or ``person walking''), by contrasting the corresponding visual tokens with the positive and negative text tokens directly.
However, this approach faces several practical limitations. The negative prompts are sampled from existing textual descriptions for a given image, which severely limits the number of plausible and useful negative samples.
As a result, the supervision signal is noisy and unstable as contrastive objectives typically require large batch sizes with numerous negative samples to learn robust fine-grained representations~\cite{khosla2020supervised, he2020moco,chen2020simclr, henaff2020cpc, tian2020cmc}. In addition, the entire contrastive process is anchored in the text embedding space, so its effectiveness strongly depends on the typically poor quality of image-text alignment and the precision of the underlying language representations. This misalignment results into visual embeddings that entangle semantically distinct but visually similar objects, limiting the model's ability to learn precise fine-grained distinctions. 

A straightforward approach to address these misalignment issues is using a contrastive objective that operates in the image space only. Even though image-space CL has been used with great effectiveness in multiple different visual tasks \cite{chen2020simclr, reid_triplet, reid_cluster_contrast, Wang_2021_ICCV}, its application in REC has not been explored. 
To this end, we introduce \textbf{C-REX} (\underline{C}ontrastive Learning for \underline{R}eferring \underline{Ex}pression Counting), a simple supervised contrastive learning framework tailored for REC that shifts the supervision from the text–image space to operate entirely in the image embedding space. Instead of relying on linguistic anchors, C-REX guides visual tokens of objects corresponding to the same RE to be embedded close together and away from the embeddings of objects corresponding to different expressions or classes, as shown on the right panel of Figure~\ref{fig:method}. This promotes a visual representation space sensitive to fine-grained visual differences that correspond to variation in semantic context. The proposed contrastive objective is more stable as the scarcity of negative samples has been addressed; the model can leverage a large pool of negatives (>100) in the form of visual tokens instead of relying on the limited REs (typically less than 5).

C-REX introduces an intra-image contrastive learning objective that operates entirely in the visual embedding space. Concretely, for each Image-RE pair we match predicted points to the $N$ ground truth points by performing Bipartite Matching. The image tokens responsible for the matched predictions are treated as positive samples in the contrastive objective, whereas all other tokens are treated as negatives. This way, the RE is only used as input to the visual backbone to condition the generated image tokens through cross-attention. Note that, unlike the standard supervised contrastive objective~\cite{khosla2020supervised}, we do not pull negative samples together, since they lack semantic coherence, but only push them away from the positives. This improves performance as shown in Section~\ref{sec:abl_scl}. 
The simple shift to image-space CL effectively sharpens the visual feature space, improving intra-class separability and enhancing the model’s ability to handle subtle, attribute-level distinctions in open-world counting.

A key strength of our framework lies in its \textbf{generality} and \textbf{versatility}. C-REX can be plugged into any REC architecture without modification. We validate this by integrating it into GroundingREC~\cite{groundingrec}, CADGD~\cite{cadgd}, and a detection-based baseline built on GroundingDINO~\cite{liu2023grounding}. For GroundingREC and CADGD, we replace their image-text contrastive loss with our image-space objective; for GroundingDINO, we add it alongside the standard counting loss. In all cases, C-REX yields consistent performance gains while preserving the original design. We further verify that combining image-text and image-space contrastive objectives does not improve performance, with image-space supervision alone consistently yielding the best results (Section~\ref{sec:abl_scl}). Beyond REC, the same principle of learning fine-grained discriminative features through contrastive alignment in the image-space only, can be applied to other counting settings. We adapt C-REX to \textbf{class-agnostic counting} by matching predicted and ground-truth points per class to form positives and treating the remaining samples as negatives. With this simple modification, C-REX also boosts performance for text-based counting models in the class-agnostic setting (Section~\ref{sec:exp}).
To summarize, our contributions are the following:
\begin{itemize}
    \item We demonstrate that counting tasks can benefit from performing contrastive learning in the image (visual) embedding space \emph{only}, rather than in a typically poorly aligned image-text space. We introduce C-REX, a supervised CL framework for REC that operates directly in the visual embedding space.
    \item Our framework produces fine-grained, context-aware visual representations and achieves state-of-the-art performance across multiple REC architectures, improving by up to 28\% in MAE and 24.5\% in RMSE.
    \item We show that C-REX can be applied to any existing REC architecture and can also
    be deployed in other counting settings that require learning fine-grained discriminative representations, such as class-agnostic counting.
    
\end{itemize}
\section{Related Work}
\label{sec:rel}
Early works in object counting focused mostly on class-specific counting \cite{crowd-loc, wang2020DMCount, zhang-crowd-counting}, addressing challenges in diverse domains such as crowd counting \cite{app-crowd-counting, crowd-counting-2, app-crowd-counting-2, yang2025taste, Meng_2025_CVPR} or traffic analysis \cite{app-traffic}. More recent work has moved to class-agnostic counting (CAC)~\cite{zhizhong2024point, tfoc, fsc-147} where the aim is to generalize counting across various object categories, typically not encountered during training \cite{class-agnostic, class-agnostic-2, class-agnostic-3, Xu2023LearningFP, multi-class-2, dumery2025counting}. To enable this, many methods used textual description to describe the classes \cite{vlcounter, countr, countx, amini2025open, zhang2025enhancing, Liu_2025_ICCV}, while others use a small number of annotated examples as visual reference \cite{few-shot-1, few-shot-2, diffcount, huang2024count, chen2025single} to achieve better results. A further line of work investigates whether the reasoning capabilities of Vision–Language Models (VLMs) can also enhance performance~\cite{pothiraj2025capture, alghisi2025re}. VLMs are rapidly improving, however in practice, they increasingly rely on external, task-specific components to use as tools \cite{schick2023toolformer}. Counting models specifically have been utilized in this manner \cite{amini2025countgd++} and have shown to outperform general use VLMs, outlining the continuing need for dedicated counting models.

Another way the counting task has been approached is in the context of Visual Question Answering (Counting VQA) \cite{vqa1, vqa2, tallyqa, countqa, howmanyqa}, where a VQA model is tasked with answering a question that requires counting objects in an image. TallyQA\cite{tallyqa} focuses on complex counting scenarios where the VQA model has to distinguish between visually similar objects given fine-grained textual descriptions. Similarly, Dai \etal \cite{groundingrec} introduced REC, where the goal is to count only the subsets of instances of a class in an image that match a given RE. 
Each RE consists of a \textbf{class}, specifying the object category, and an \textbf{attribute}, describing its fine-grained characteristics. Direct counting methods like REC differ from counting VQA in that the models are tasked to also localize the predicted objects to ensure more trustworthy predictions.
REC is also closely related to Referring Expression Segmentation \cite{res-1, res-2, res-3, luo2025cohd} and Referring Expression Comprehension \cite{re-comp-1, re-comp-2}, which focus on segmentation or localization rather than counting.

Early approaches on object counting mostly relied
on density map regression \cite{arteta-density, lempitsky-density-2, density-3} or bounding box detection \cite{mask-rcnn, trad-iv} in order to predict the object count. Density-based methods estimate the final count by summing over a density map, being traditionally more accurate in cluttered and densely populated scenes where detection-based methods struggle \cite{countgd}. However, a critical limitation of density-based methods is the lack of object-level detail which limits their applicability in cases that require localization and explainability \cite{density-issues, density-issues-2}. This is especially important in REC, where images contain multiple visually similar instances of the same category, each associated with a different referring expression. In such cases, ensuring that the final count accurately reflects only the correct subset of objects is crucial for reliable and precise counting. Recent approaches, such as CountGD \cite{countgd} and GeCo \cite{geco}, leverage state-of-the-art open-set object detectors \cite{zhang2023dino, liu2023grounding}, and reach performance on-par with the best density-based methods \cite{dave}. CAD-GD \cite{cadgd} combines density prediction with localization for REC establishing the state-of-the-art for this task.

For REC it is crucial to be able to disambiguate between visually similar objects with subtle contextual differences. For this reason, Dai \etal \cite{groundingrec} introduce a CL module that contrasts image and RE embeddings, which CAD-GD \cite{cadgd} also employs. Contrastive learning in the image-text space for REC has several disadvantages regarding robustness of learned representations and efficiency (see discussion in Section \ref{sec:advantages}) which we address by moving the contrastive supervision to the image space only.
CL in the image space only, beyond self-supervised pretraining \cite{he2020moco,chen2020simclr, henaff2020cpc, tian2020cmc}, has also been explored and seen success in other vision tasks such as person re-identification (ReID) \cite{reid_triplet, reid_triplet_again, reid_cluster_contrast, reid_unify}, where the task is to distinguish between people that are potentially visually similar but correspond to distinct identities. 
REC faces a similar challenge: models must distinguish between visually similar objects (same class), while discriminating based on the RE context. Our approach in REC differs from those in ReID as the CL samples are visual tokens originating from the same image, instead of different images. Additionally, we define positive and negative sets for CL on the fly by matching predicted image tokens to ground truth targets conditioned on each RE.

\section{Method}
\label{sec:method}

For REC it is important to be able to robustly differentiate objects that belong in the same class but have different attributes (REs). To this end we propose a novel contrastive learning module, that learns more robust discriminative features using contrastive supervision on the image space only. 

Our goal is to perform contrastive learning in the image embedding space only, as text-image contrastive learning tends to be less stable. In the context of REC, we want our approach to accommodate multiple positive samples, since an image typically contains multiple instances of objects that correspond to the same class-RE pair. To ensure effective representation learning, we aim to bring the embeddings of these instances closer together while maintaining clear separation from objects that correspond to the same class but different RE or a different class altogether. For this reason our proposed loss is based on the supervised contrastive loss \cite{khosla2020supervised} that extends the standard contrastive loss to be able to handle multiple positive samples for each anchor. The general formulation of the supervised contrastive objective is as follows: Given a set $I$ of samples (\eg images), for each sample $i \in I$ we get the embeddings $z_i$, and a set of other samples that correspond to the same ``label'' as sample $i$, which we call the set of positives $p\in P(i)$ for $i$. The loss is calculated as:

\begingroup
\small
\begin{align}
\mathcal{L}_{\sup} = \sum_{i \in I}
  -\log\left(
    \frac{1}{\lvert P(i)\rvert}
    \sum_{p \in P(i)}
      \frac{\exp(z_i \cdot z_p \,/\, \tau)}
           {\sum\limits_{\alpha \in I - \{i\}}
             \exp(z_i \cdot z_\alpha \,/\, \tau)}
  \right)
\end{align}
\endgroup

For our approach, we choose the samples $i\in I$ to correspond to predicted image tokens from our visual backbone. We use the generated image tokens to predict the location of the target objects, as is standard in counting models \cite{groundingrec, cadgd, countgd}. We aim to contrast the image tokens that correspond to the objects for a given class and RE (same attributes, semantic context) to the ones that correspond to the same class but different RE or to different classes altogether. To assign labels to our samples $i \in I$ we match them to the $N$ ground truth points via Bipartite Matching, based on their predicted locations. For each Image-RE pair we divide the samples into a positive ($I^+$) and negative ($I^-$) class; the image tokens that get matched to ground truth points correspond to the positive class and all other tokens are treated as negatives.

The standard formulation of the supervised contrastive loss pushes samples from the same class closer together. In our case this would mean that samples from the negative class $I^-$, which may contain samples from a diverse set of class labels, REs, or even tokens corresponding to no class label would be clustered together, which is undesirable and counterproductive. For example, for the image in Figure \ref{fig:method}, given the RE ``Person riding a bicycle'', the set of negatives would include tokens that correspond to ``people walking'', ``cars'' or ``traffic lights'', which should not be clustered together. To address this, we modify the supervised contrastive objective to only use samples from the positive class as anchors. Based on this modification, our revised supervised contrastive loss can be formulated as follows:

\begingroup
\begin{align}
\mathcal{L}^{*}_{\sup}=
\underbrace{\sum_{i \in I^+}}_{\substack{\textbf{optimize}\\ \textbf{positives } \textbf{only}}}
  -\log\!\left(\frac{}{}
    \frac{1}{\lvert I^+\rvert-1}
    \sum_{p \in I^+-\{i\}}
      \frac{\exp(z_i \cdot z_p \,/\, \tau)}
           {\sum\limits_{\alpha \in I-\{i\}}
             \exp(z_i \cdot z_\alpha \,/\, \tau)}
  \right)
\end{align}
\endgroup

We verify that the motivation for this modification is well-grounded and supported by demonstrating its effectiveness in our ablation study (Section \ref{sec:abl}).

Our final framework, \textbf{C-REX}, replaces the image–text contrastive loss used in previous REC methods \cite{groundingrec, cadgd} with our proposed image-space contrastive loss and uses all other counting losses (\eg localization, classification or density loss) unchanged for each model we apply it to.

\subsection{Advantages of image-space over image-text CL methods}
\label{sec:advantages}
Previous work on REC \cite{groundingrec, cadgd} also tried to learn more discriminative features by contrasting image embeddings to text embeddings of ``candidate'' REs. However, their approach has key limitations in the context of contrastive learning. Specifically, their contrastive loss operates on image-text pairs, which presents additional challenges due to the inherent misalignment between image and text embedding spaces \cite{li2021albef}. In contrast, our method operates entirely within the image space, ensuring a more stable contrastive signal. Contrastive learning in a language-image setting generally requires more data and larger-scale training to achieve robust representations \cite{clip, cherti2023openclip}, whereas contrastive learning in the image space is more efficient \cite{zhang2023dino, oquab2024dinov2}. This is evident when comparing the performance of text-image models like CLIP \cite{clip} to image-specific models like DINOv2 \cite{oquab2024dinov2} on smaller datasets, where DINOv2 achieves comparable performance to CLIP even when trained on significantly less data \cite{oquab2024dinov2}. 

Additionally, a crucial factor in contrastive learning is the presence of a large batch size with numerous negative samples, as this increases the diversity and variability of samples included, thus improving the robustness of the learned representations \cite{khosla2020supervised, he2020moco,chen2020simclr, henaff2020cpc, tian2020cmc}. However, \cite{groundingrec, cadgd} are limited in that regard, as the number of negative samples is constrained by the number of existing REs for a single class in an image, typically fewer than four. Our contrastive loss overcomes this constraint by leveraging a significantly larger pool of negatives, typically in the hundreds, consisting of image tokens that were not matched to ground truth objects, and so ended up in the negative set of our contrastive framework. These advantages are reflected in our experimental results (Section~\ref{sec:exp}), where we compare two state-of-the-art models~\cite{groundingrec, cadgd} under their original image–text contrastive framework and our proposed C-REX. In both cases, replacing the traditional contrastive objective with C-REX yields consistent improvements across all evaluation metrics.

\subsection{Extension to class-agnostic counting}
C-REX is designed to be general and versatile, enabling models to distinguish objects with subtle, context-dependent variations, such as those described by REs. This allows us to adapt it to any task that requires multiple image-space object embeddings to be contrasted based on a set of distinguishing attributes. Following this principle, we introduce a simple adaptation for class-agnostic counting,  where positive samples are selected based on a target class, rather than an RE. This is less critical in this counting setting where models need to differentiate between objects of different classes, whose image embeddings are already different enough, so we do not expect to see the same amount of improvement as in REC. With this adaptation, our method can improve the performance of the best text-based counting models in the class-agnostic counting task, as shown in Section \ref{sec:exp}, although the improvement is less significant, as was expected.

\section{Experiments}
\label{sec:exp}

We evaluate C-REX as a general framework that can be applied to existing REC architectures. Specifically, we integrate our proposed framework into two state-of-the-art REC models; GroundingREC \cite{groundingrec} and CAD-GD \cite{cadgd}, as well as a simple detection-based baseline built on GroundingDINO \cite{liu2023grounding}. For the former two, we replaced their original image-text contrastive objectives with our image-space contrastive loss. For the GroundingDINO baseline, we train using our proposed objective alongside the standard counting objectives used in previous detection-based models \cite{countgd, groundingrec, cadgd}.

\begin{table}[h!]
\centering
\caption{Comparison of results on the REC-8K dataset. GrREC and GrDINO correspond to GroundingREC~\cite{groundingrec} and GroundingDINO~\cite{liu2023grounding}. For GroundingREC~\cite{groundingrec}, results are obtained from their released models. For CAD-GD~\cite{cadgd}, since no pre-trained models are available, we retrain on the REC-8K dataset using their publicly available code. CAD-GD$\dagger$ refers to an alternative strategy of selecting positive detections, described by \cite{cadgd}, where instead of using a static threshold for all tokens, we select the top $N$ based on the density map estimate. Best result for each backbone in bold.}
\begin{tabular}{V l ccccc ccccc}
\toprule
\multirow{2}{*}{} & \multirow{2}{*}{Method}
& \multicolumn{5}{c}{Validation set}
& \multicolumn{5}{c}{Test set} \\
\cmidrule(lr){3-7} \cmidrule(lr){8-12}
& & MAE$\downarrow$ & RMSE$\downarrow$ & Prec$\uparrow$ & Rec$\uparrow$ & F1$\uparrow$
  & MAE$\downarrow$ & RMSE$\downarrow$ & Prec$\uparrow$ & Rec$\uparrow$ & F1$\uparrow$ \\
\midrule

& GrREC           & 6.80  & 18.13 & 0.65 & 0.71 & 0.68 & 6.50  & 19.79 & 0.67 & 0.72 & 0.69 \\
\rowcolor{darkgray!10} \cellcolor{white}
& \enspace + C-REX      & 4.74  & 14.50 & 0.70 & 0.71 & 0.71 & 4.91  & 18.87 & 0.71 & 0.72 & 0.72 \\
\cdashline{2-12}
\addlinespace[1.5pt]

& CAD-GD         & 6.33  & 15.90 & 0.65 & 0.69 & 0.67 & 6.25  & 18.91 & 0.67 & 0.70 & 0.69 \\
\rowcolor{darkgray!10} \cellcolor{white}
& \enspace + C-REX      & 5.98  & 14.84 & 0.66 & 0.70 & 0.68 & 5.97  & 17.37 & 0.67 & 0.73 & 0.70 \\
\cdashline{2-12}
\addlinespace[1.5pt]

& CAD-GD$^\dagger$      & 4.65  & 12.87 & 0.65 & 0.69 & 0.67 & 4.81  & 15.02 & 0.67 & 0.70 & 0.69 \\
\rowcolor{darkgray!10} \cellcolor{white}
& \enspace + C-REX      & 4.70  & 13.78 & 0.66 & 0.70 & 0.68 & 4.61  & \textbf{14.71} & 0.67 & 0.73 & 0.70 \\
\cdashline{2-12}
\addlinespace[1.5pt]

& GrDINO         & 5.92  & 17.09 & 0.65 & 0.65 & 0.65 & 5.90  & 19.73 & 0.68 & 0.68 & 0.68 \\
\rowcolor{darkgray!10} \cellcolor{white}
\multirow{-8}{=}{\rotatebox{90}{\large\strut \textbf{Swin-T}}}
& \enspace + C-REX      & \textbf{4.26} & \textbf{12.89} & \textbf{0.75} & \textbf{0.72} & \textbf{0.73}
                         & \textbf{4.38} & 17.93 & \textbf{0.76} & \textbf{0.74} & \textbf{0.75} \\
\midrule

& GrREC           & 5.66  & 15.24 & 0.66 & 0.77 & 0.71 & 5.42  & 18.47 & 0.71 & 0.69 & 0.70 \\
\rowcolor{darkgray!10} \cellcolor{white}
& \enspace + C-REX      & 4.57  & 14.22 & 0.72 & 0.72 & 0.72 & 4.58  & 18.07 & 0.74 & 0.72 & 0.73 \\
\cdashline{2-12}
\addlinespace[1.5pt]

& CAD-GD         & 5.31  & 13.68 & 0.70 & 0.73 & 0.72 & 5.36  & 16.14 & 0.73 & 0.73 & 0.73 \\
\rowcolor{darkgray!10} \cellcolor{white}
& \enspace + C-REX      & 5.06  & 12.73 & 0.70 & 0.74 & 0.72 & 5.02  & 15.09 & 0.72 & 0.74 & 0.73 \\
\cdashline{2-12}
\addlinespace[1.5pt]

& CAD-GD$^\dagger$      & 4.40  & \textbf{12.20} & 0.70 & 0.73 & 0.72 & 4.58 & 13.81 & 0.73 & 0.73 & 0.73 \\
\rowcolor{darkgray!10} \cellcolor{white}
& \enspace + C-REX      & 4.38  & 12.43 & 0.70 & 0.74 & 0.72 & 4.49  & \textbf{13.21} & 0.72 & 0.74 & 0.73 \\
\cdashline{2-12}
\addlinespace[1.5pt]

& GrDINO         & 4.58  & 14.64 & 0.70 & 0.76 & 0.73 & 4.64  & 17.58 & 0.72 & 0.76 & 0.74 \\
\rowcolor{darkgray!10} \cellcolor{white}
\multirow{-8}{=}{\rotatebox{90}{\large\strut \textbf{Swin-B}}}
& \enspace + C-REX      & \textbf{4.00} & 14.49 & \textbf{0.78} & \textbf{0.76} & \textbf{0.77}
                         & \textbf{4.20} & 16.76 & \textbf{0.79} & \textbf{0.77} & \textbf{0.78} \\

\bottomrule
\end{tabular}
\label{tab:rec8k}
\end{table}

\subsection{Implementation Details}
\label{subsec:impl-details}
For GroundingREC \cite{groundingrec}, we use the publicly released model checkpoint and evaluation protocol. Since the model and its configuration were reproducible, we directly report their results as our baseline performance. For CAD-GD~\cite{cadgd}, since no trained checkpoint is publicly available, we retrained the model using the authors’ official open-source implementation. For the required density maps, we followed the same procedure as the authors, established in prior work~\cite{lempitsky-density-2}. More details are provided in the Supplementary.

\smallskip
\noindent\textbf{Training} For all models, we use either a Swin-T \cite{swin} or Swin-B image encoder and BERT \cite{devlin2019bert} text encoder, both kept frozen during training. We fine-tune only the feature enhancer and cross-modality decoder components. For CAD-GD~\cite{cadgd}, we follow their training procedure. For the REC-8k dataset we train for 20 epochs and use a learning rate of 1e-5 that decays by 10x in the 10th epoch and for FSC-147 we train for 30 epochs and use a learning rate of 1e-5 that decays by 10x in the 15th epoch. For GroundingREC and GroundingDINO we follow Dai \etal \cite{groundingrec} and do not decay the learning rate during training.

\smallskip
\noindent\textbf{Inference} Following prior works \cite{detr, groundingrec, cadgd}, we generate $900$ output tokens per image and apply the thresholding strategy of Dai \etal~\cite{groundingrec} (0.25 for the CLS token, 0.35 for the remaining tokens) to determine positive detections. For CAD-GD we also evaluate using their alternate proposed strategy (CAD-GD$\dagger$), where, instead of using a static threshold, we select the top $N$ candidate tokens based on the density map estimate as positive detections.

\subsection{Datasets \& Metrics}
\label{subsec:datasets-metrics}
\smallskip
\noindent\textbf{REC-8K} \cite{groundingrec} consists of $8,011$ images, each paired with $2.13$ REs on average. In total, the dataset contains $17,122$ image-RE pairs, where each pair is annotated with the ground-truth points pinpointing the target objects in the image.

\smallskip
\noindent\textbf{FSC-147} \cite{fsc-147} is a widely used object counting dataset, which consists of $6,135$ images spanning $147$ classes, split in a non-overlapping manner. Each image is annotated with three visual exemplars.

\smallskip
\noindent\textbf{Metrics} Following prior work \cite{Xu2023ZeroShotOC, countgd, dave, groundingrec}, we report the Mean Absolute Error (MAE) and Root Mean Squared Error (RMSE) to evaluate and compare counting performance. Moreover, we also report precision, recall and F1 score since counting the correct object instances is also an important factor.

\begin{figure}[h!]
  \centering
  \includegraphics[width=\textwidth]{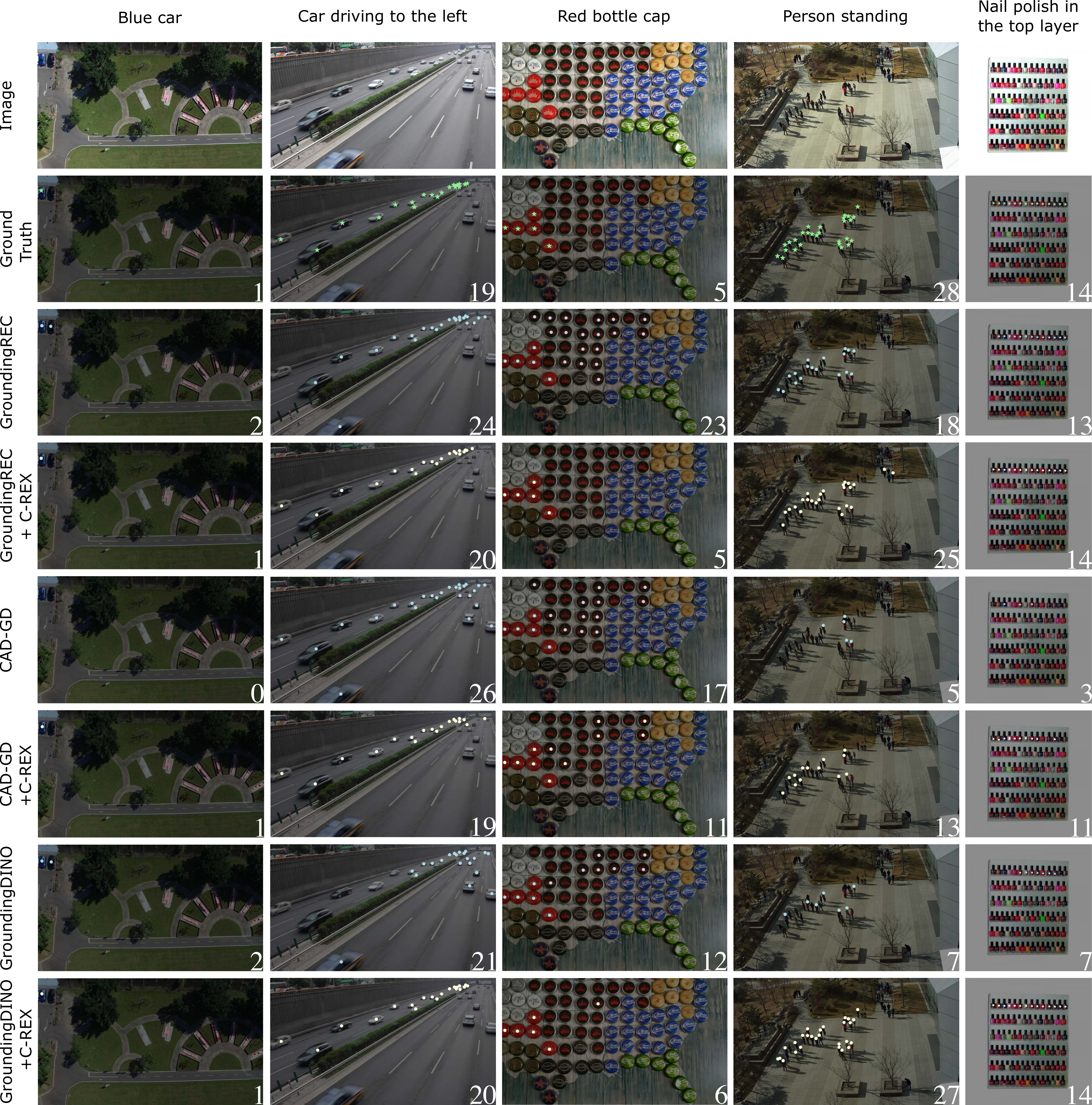}
  \caption{Qualitative comparison of REC predictions across different models. The first row shows the input images and the second row visualizes the ground-truth instance annotations for each referring expression. The remaining rows present predictions from each baseline model (GroundingREC, CAD-GD, and GroundingDINO) followed by their corresponding versions with C-REX applied. Each column corresponds to a different referring expression. Across all examples, models trained with C-REX provide more precise counts and localize a more accurate subset of objects specified by the referring expression. Image contrast is adjusted to better visualize model predictions.}
  \label{fig:qual}
\end{figure}

\begin{figure}[t!]
  \centering
  \includegraphics[width=0.95\textwidth]{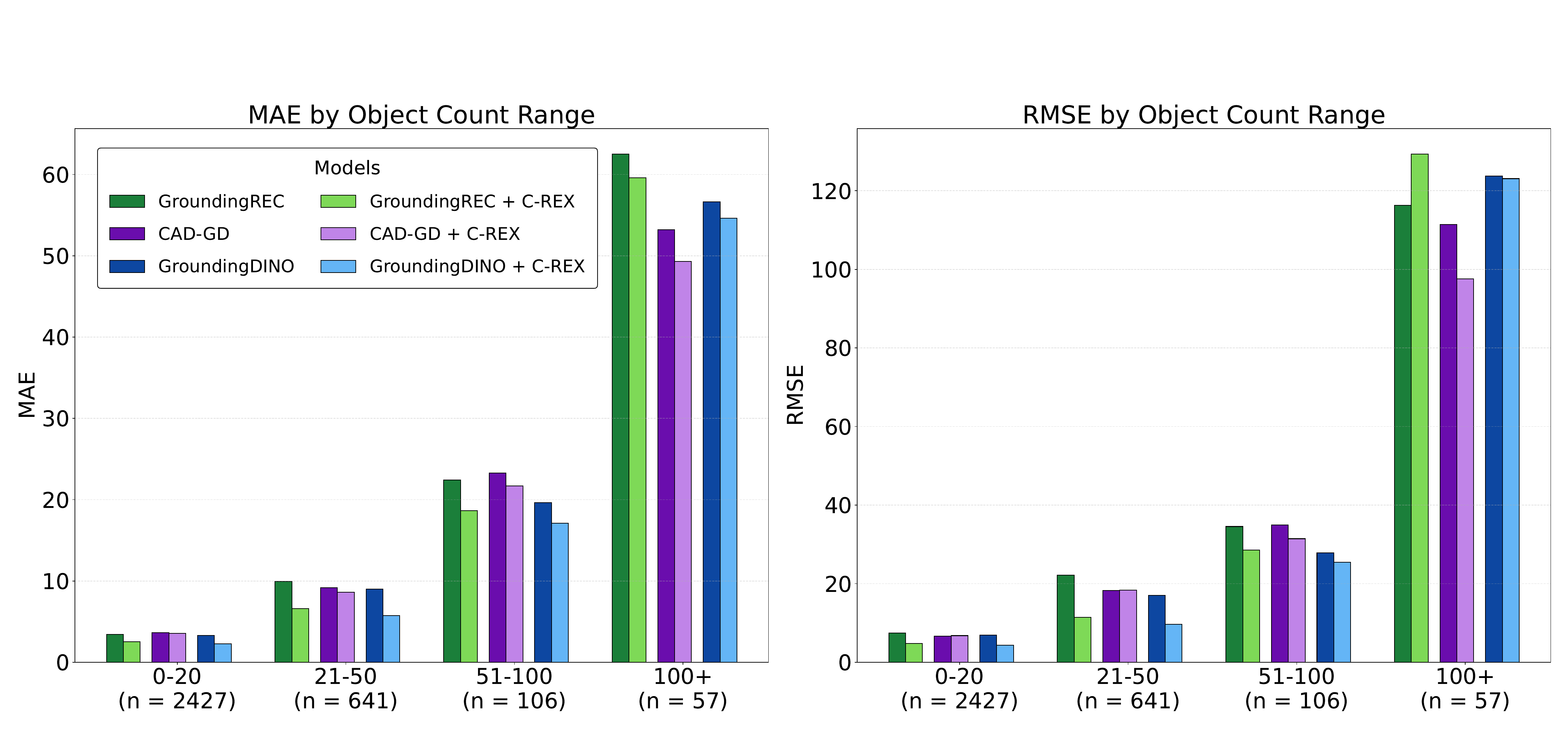}
  \caption{Performance of each baseline REC model compared to its counterpart trained with our proposed C-REX framework, evaluated across object count ranges on REC-8K. Sample counts are below each bin. In most cases, C-REX yields lower MAE and RMSE compared to the corresponding baseline configuration, demonstrating consistent improvements across low, mid, and high-density conditions.}
  \label{fig:histo}
\end{figure}

\subsection{Results in REC}
\label{subsec:results}

\smallskip
\noindent\textbf{Quantitative Comparison} 
Table~\ref{tab:rec8k} presents our experimental results in the REC-8K dataset, comparing the performance of three detection-based REC architectures--GroundingREC \cite{groundingrec}, CAD-GD \cite{cadgd}, and GroundingDINO \cite{liu2023grounding}--evaluated both in their original configurations and with C-REX applied. All architectures are evaluated using Swin-T and Swin-B backbones. As shown in Table~\ref{tab:rec8k}, integrating C-REX consistently improves performance across models, backbones and metrics, confirming that replacing the image–text contrastive loss with our image-space objective yields more stable and discriminative representations. C-REX achieves the lowest overall errors and the highest F1 scores (results in \textbf{bold}) in most settings, establishing new state-of-the-art performance on REC-8K. We also find that using just C-REX improves performance compared to using both image-text and image-space contrastive losses jointly (Section \ref{sec:abl}).

Figure~\ref{fig:histo} reports MAE and RMSE across different object count ranges on the REC-8K test set, allowing us to examine model behavior under varying object densities. Across all three REC architectures, integrating C-REX consistently reduces error relative to the corresponding baseline across all count regimes. The improvements are particularly evident in the low- and mid-density ranges (0–20, 21–50, and 51–100 objects), where fine-grained attribute distinctions make the task particularly challenging. These results indicate that the proposed image-space contrastive supervision improves fine-grained instance discrimination and leads to more accurate counting across a wide range of object densities.

C-REX forms positive and negative sets from the fixed number of tokens provided by the visual backbone, so extremely dense scenes can reduce the number of tokens available to serve as negatives. While this could lead to a lower quality supervision signal, we do not observe a negative effect in model performance. Even in high-count regimes, C-REX usually provides a sufficient number of negative image tokens. 
Additionally, as vision backbone models advance and continue to increase their visual token capacity, the available pool of negatives will organically grow, allowing C-REX to scale more effectively with more expressive backbones than image-text CL methods.

\begin{table}[t]
\small
\centering
\renewcommand{\arraystretch}{1.1}
\caption{Comparison of results on the FSC-147 \cite{fsc-147} dataset, using only textual information (class names) as a guide to perform counting. For all models, except CAD-GD and GroundingDino, we report results from their original papers. Best results in bold.}
\small
\begin{tabular}{l c c c c}
\toprule
\multirow{2}{*}{Method}  & \multicolumn{2}{c}{Val set} & \multicolumn{2}{c}{Test set} \\
\cmidrule(lr){2-3} \cmidrule(lr){4-5}
 & MAE & RMSE & MAE & RMSE \\
\midrule
ZSC \cite{Xu2023ZeroShotOC}  & 26.93 & 88.63 & 22.09 & 115.17 \\
CounTX \cite{countx}  & 17.10 & 65.61 & 15.88 & 106.29 \\
CountGD \cite{countgd}& 12.14 & \textbf{47.51} & 14.76 & 120.42\\
DAVE \cite{dave}  & 15.48 & 52.57 & 14.90  & 103.42\\
\midrule
CAD-GD \cite{cadgd} & 13.80 & 64.74 & 12.15  & 99.87\\
\rowcolor{darkgray!10}
\quad + C-REX & 13.86 & 61.04 & 11.90  & \textbf{99.65}\\
GroundingDino & 9.71  & 55.11 &10.73  &103.79 \\
\rowcolor{darkgray!10}
\quad + C-REX & \textbf{9.42} & 54.74 & \textbf{10.04} & 102.35 \\
GroundingREC \cite{groundingrec}  & 10.06 & 58.62 & 10.12 & 107.19 \\
\rowcolor{darkgray!10}
\quad + C-REX  & 9.66 & 54.07 & 12.46 & 104.53 \\
\bottomrule
\end{tabular}
\label{tab:fsc147}
\end{table}

\smallskip
\noindent\textbf{Qualitative Comparison}
Figure~\ref{fig:qual} presents qualitative comparisons across the three REC architectures both in their original configurations and with C-REX applied. Each column corresponds to a different RE, covering a diverse set of fine-grained distinctions such as object color (``blue car''), orientation (``car driving to the left''), positional relationships (``nail polish in the top layer''), and actions (``person standing''). The first two rows show the input images and ground-truth annotations, while the remaining rows present predictions from each REC model, both in their original baseline form and with C-REX applied. 

Across all cases, we observe that integrating C-REX leads to more accurate predictions, with regards to both predicted object count and locations. For instance, in the ``car driving to the left'' example, all three baseline models overcount by including vehicles moving in the opposite direction, whereas their C-REX counterparts more consistently focus on the intended subset. The ``blue car'' example in the first column also highlights the effect of C-REX: despite the scene containing only two vehicles, all baseline models fail to isolate the correct car, either detecting both or missing it entirely. The distinction between the blue car and the darker neighboring one is subtle yet semantically important, and C-REX consistently identifies the correct one. 
Moreover, in the ``nail polish in the top layer'' example, C-REX demonstrates better spatial discriminative ability by concentrating predictions to the correct layer. 
We provide more examples of both well performing and failure cases in the supplementary.

\subsection{Results in Class-Agnostic Counting}
\label{subsec:cac-results}
To assess the generality of C-REX beyond REC, we adapt it for the class-agnostic counting setting using the FSC-147 benchmark dataset~\cite{fsc-147}. We apply C-REX to our three detection-based baseline architectures ~\cite{groundingrec, cadgd, liu2023grounding} and compare each model’s performance in both its original (vanilla) configuration and with C-REX integrated. Table~\ref{tab:fsc147} reports the results, including comparisons with state-of-the-art class-agnostic counting models. In most cases, incorporating C-REX yields consistent improvements across all architectures and both evaluation metrics, indicating that C-REX transfers effectively to class-agnostic counting and benefits performance independent of the underlying model design.

\begin{table}[t]
\centering
{\small
\caption{Comparison of contrastive learning strategies: image–text {(I-T)}, image–space \textit{(I, \textbf{ours})}, and their combination {(I-T \& I)}. Across two architectures, image-space contrastive learning consistently outperforms the standard image–text objective, while combining the two does not yield further improvements. This suggests that contrastive supervision in the image space alone provides the most effective signal for REC.}
\begin{tabular}{lccccccc}
\toprule
\multirow{2}{*}{Method} & \multirow{2}{*}{CL} 
& \multicolumn{3}{c}{Val set} 
& \multicolumn{3}{c}{Test set} \\
\cmidrule(lr){3-5} \cmidrule(lr){6-8}
& & MAE$\downarrow$ & RMSE$\downarrow$ & F1$\uparrow$
  & MAE$\downarrow$ & RMSE$\downarrow$ & F1$\uparrow$ \\
\midrule

\multirow{3}{*}{GroundingREC        }  
& I-T        & 6.80 & 18.13 & 0.68 & 6.50 & 19.79 & 0.69 \\
& I-T \& I  &   5.39   &  15.39   &   0.68   &   5.36   &   19.17    &   0.69   \\
& I        & \textbf{4.74} & \textbf{14.50} & \textbf{0.71} & \textbf{4.91} & \textbf{18.87} & \textbf{0.72} \\
\midrule
\multirow{3}{*}{CAD-GD}   
& I-T        & 6.33 & 15.90 & 0.67 & 6.25 & 18.91 & 0.69 \\
& I-T \& I  &   6.15   &  \textbf{14.56} &   0.67   &  6.47  &    19.04   &   0.68   \\
& I       & \textbf{5.98} & 14.84 & \textbf{0.68} & \textbf{5.97} & \textbf{17.37} & \textbf{0.70} \\
\bottomrule
\end{tabular}

\label{tab:abl_img}}
\end{table}

\subsection{Ablation Study}
\subsubsection{Image-Text vs Image-Space CL} We compare the image-text contrastive loss used in previous REC methods \cite{groundingrec, cadgd} to our proposed image-image CL and explore whether they can be complementary (Table~\ref{tab:abl_img}). Image–space CL consistently outperforms image–text CL, as well as their combination, indicating that contrastive supervision in the image space is more effective for REC. Adding image–space CL on top of image–text CL still improves over the image–text baseline, further supporting that image-text CL is suboptimal for REC.
\label{sec:abl}

\smallskip
\noindent\textbf{Selection of positive samples}
We verify that choosing the positive set by Bipartite Matching outperforms alternative potential selection strategies. We compare it to strategies that select the most similar tokens compared to the text prompt. We show this for different thresholds for the number of most similar tokens: a static threshold, the ground truth count $N$ and $N$ + a small dynamic window to account for the uncertainty of this selection strategy (Table~\ref{tab:ablation_N_ablation_loss}a). Our matching-based selection strategy outperforms alternatives.

\begin{table}[h!]
    \small
    \centering
    \caption{\textbf{(a)} Ablation on positive sample selection strategy in our proposed contrastive framework. $N$ is the ground truth object count and \textit{BM} stands for Bipartite Matching. \textbf{(b)} Ablation on the GroundingDino (GrDino) baseline comparing our modified supervised contrastive loss ($\mathcal{L}^*_{\text{sup}}$) with the standard formulation ($\mathcal{L}_{\text{sup}}$) on REC-8k. The proposed modification consistently improves performance across metrics. For this analysis, we select as positives the top-$N$ image tokens most similar to the text prompt.}
    
\begin{subtable}[b]{0.48\linewidth}
\small
\centering
\begin{tabular}{l| cc|cc}
\toprule
\multirow{2}{*}{\textbf{\shortstack{Selection\\Strategy}}} & \multicolumn{2}{c|}{\textbf{Val set}} & \multicolumn{2}{c}{\textbf{Test set}} \\
& MAE & RMSE & MAE & RMSE  \\
\hline
top 5 & 7.58 & 18.34 &  7.51 & 20.83 \\
top $N+\sqrt{N}$ & 6.02 & 15.79 & 6.07 & 18.80 \\
top $N$& 4.86 & 13.60 & 5.06 & \textbf{17.53} \\
BM & \textbf{4.26} & \textbf{12.89} & \textbf{4.38} & 17.93 \\
\bottomrule
\end{tabular}
\caption{}
\end{subtable}
\hfill
\begin{subtable}[b]{0.46\linewidth}
\small
\centering
\begin{tabular}{l| cc|cc}
\toprule
\textbf{Method} & \multicolumn{2}{c|}{\textbf{Val set}} & \multicolumn{2}{c}{\textbf{Test set}} \\
& MAE & RMSE & MAE & RMSE \\
\hline
GrDino & 5.92 & 17.09 & 5.90 & 19.73 \\
+ $\mathcal{L}_{\text{sup}}$ & 5.69 & 14.98 & 5.91 & 19.34 \\
+ $\mathcal{L}^*_{\text{sup}}$ & \textbf{4.86} & \textbf{13.60} &
\textbf{5.06} & \textbf{17.53} \\
\bottomrule

\end{tabular}
\caption{}
\end{subtable}
    \label{tab:ablation_N_ablation_loss}
\end{table}

\smallskip
\noindent\textbf{Supervised CL modification}
\label{sec:abl_scl}
In Table \ref{tab:ablation_N_ablation_loss}b we examine the effectiveness of our proposed modification ($\mathcal{L}^*_{\text{sup}}$) to the supervised contrastive objective. We show that it significantly improves performance compared to both the unmodified version and the vanilla configuration of our GroundingDino baseline. This highlights the importance of our decision to use \textbf{only positive} samples as anchors, confirming its crucial role in improving the model’s effectiveness.

\begin{table}[h!]
    \small
    \centering
    \caption{\textbf{(a)} Robustness to paraphrased REs on the REC-8K test set. C-REX improves performance for two of the three REC architectures, indicating robustness to linguistic variations. \textbf{(b)} Comparison of Molmo2-4B~\cite{deitke2025molmo} on the REC-8K test set against REC models equipped with C-REX. Molmo performs substantially worse, highlighting the importance of specialized visual representations for REC. }
    \begin{subtable}[t]{0.46\linewidth}
\centering
\small
\resizebox{\linewidth}{!}{
\begin{tabular}{lccc}
\toprule
\multirow{2}{*}{Method} & \multicolumn{3}{c}{Test set} \\
\cmidrule(lr){2-4}
& MAE$\downarrow$ & RMSE$\downarrow$ & F1$\uparrow$ \\
\midrule
GrREC & 9.88 & 24.77 & 0.54 \\
GrREC + C-REX & \underline{6.22} & \underline{19.41} & \underline{0.66} \\
\midrule
CAD-GD & \underline{6.85} & \underline{18.29} & \underline{0.68} \\
CAD-GD + C-REX & 8.02 & 20.86 & 0.67 \\
\midrule
GrDINO & 6.34 & 21.35 & 0.67 \\
GrDINO + C-REX & \underline{5.66} & \underline{19.48} & \underline{0.72} \\
\bottomrule
\end{tabular}
}
\caption{}
\label{tab:simplified_test}
\end{subtable}
\hfill
\begin{subtable}[t]{0.5\linewidth}
\centering
\small
\resizebox{\linewidth}{!}{
\begin{tabular}{lccc}
\toprule
\multirow{2}{*}{Method} & \multicolumn{3}{c}{Test set} \\
\cmidrule(lr){2-4}
& MAE$\downarrow$ & RMSE$\downarrow$ & F1$\uparrow$ \\
\midrule
Molmo 0-shot & 10.04 & 32.18 & 0.51 \\
Molmo few-shot ($n=2$) & 25.43 & 42.10 & 0.39 \\
\midrule
GroundingREC + C-REX & 4.58 & 18.07 & 0.73 \\
CAD-GD + C-REX & 5.02 & \textbf{15.09} & 0.73 \\
GroundingDINO + C-REX & \textbf{4.20} & 16.76 & \textbf{0.78} \\
\bottomrule
\end{tabular}
}
\caption{}
\label{tab:paraphrase}
\end{subtable}
\label{tab:additional_results}
    \label{tab:comparison_with_molmo_and_paraphrasing}
\end{table}

\smallskip
\noindent\textbf{Robustness to Referring Expression Paraphrases}
To evaluate robustness to linguistic variation, we paraphrase the REs in the REC-8K test set using Claude~\cite{anthropic2024claude3} and evaluate each model on the paraphrased expressions (Table~\ref{tab:comparison_with_molmo_and_paraphrasing}a). C-REX improves performance for GroundingREC and GroundingDINO across all metrics, suggesting that image-space contrastive supervision can improve robustness to text perturbations, while for CAD-GD, the baseline performs better.

\subsection{Comparison with off-the-shelf foundation VLM}
We evaluate Molmo2-4B~\cite{deitke2025molmo} in REC by prompting it to \textit{``Point to all {$<$RE$>$} in the image’’}, either in a zero-shot setting or in a two-shot setting on the REC-8K test set with examples selected from the same semantic category as the query. Molmo, despite being partially pre-trained on counting, performs substantially worse than all REC models equipped with C-REX (Table~\ref{tab:comparison_with_molmo_and_paraphrasing}b). Interestingly, even prompting with semantically relevant in-context examples leads to worse performance, showing that REC remains challenging for modern MLLMs.

\section{Conclusion}
\label{sec:conclusion}
In this work, we introduced C-REX, a novel contrastive learning framework designed to tackle Referring Expression Counting (REC) by improving detection-based counting models' capabilities in distinguishing visually similar but contextually distinct objects. By operating entirely within the image space, C-REX eliminates the misalignment issues inherent in image-text contrastive learning, ensuring a more stable contrastive signal and leveraging a significantly larger pool of negative samples for improved robustness. We demonstrated the effectiveness of C-REX by applying it on three existing REC architectures \cite{groundingrec, cadgd, liu2023grounding}, consistently improving their performance and achieving state-of-the-art results. Beyond REC, we showed that the same framework extends naturally to class-agnostic counting, underscoring its generality and adaptability across related visual counting tasks.

\section*{Acknowledgements}
This research was partially supported by NSF grants IIS-2123920 and IIS-2212046 and by the CCI startup fund at UNC Charlotte.
%
%
\bibliographystyle{splncs04}
\bibliography{main}

\clearpage
\setcounter{page}{1}

\noindent\textbf{\Large{\begin{center} What is the Right Embedding Space for Contrastive Learning in Referring Expression Counting?\end{center}}}

\begin{center} Supplementary Material \end{center}

\section{Evaluation Metrics}
\label{sec:eval_metrics}
We use MAE, RMSE and F1 as metrics.
Given $n$ samples where $c_i$ represents the ground truth counts and $\hat{c_i}$ represents the predicted counts for each sample  $i$, the MAE and RMSE are calculated as:

\begingroup
\small
\begin{align}
    \text{MAE} = \frac{1}{n}\sum_{i=1}^n |c_i - \hat{c_i}|, \quad \text{RMSE} = \sqrt{\frac{1}{n}\sum_{i=1}^n (c_i - \hat{c_i})^2}
\end{align}
\endgroup

In the REC task specifically, images typically contain multiple instances of a given class, each associated with different referring expressions. This setup poses a significant challenge for detection-based counting models, which may miss correct instances or generate false positives by misidentifying objects of the same class that belong to different referring expressions. When this occurs, MAE and RMSE can be misleadingly low, failing to capture the true counting performance. To address this, we also report the F1 score, following Dai~\etal~\cite{groundingrec}, which ensures that our predicted count not only matches the total number of instances but also correctly identifies them.

\section{Implementation Details}
\label{sec:impl-details}
We create density maps based on the algorithm provided by Lempitsky~\etal~\cite{lempitsky-density-2}: First we assign a value to each pixel based on the number of annotated objects that fall inside the pixel region. Then we convolve the initial discrete density map with a fixed-size Gaussian kernel of size as described by Wang~\etal~\cite{cadgd}. Finally, we normalize the density map.

\begin{table}[!t]
\centering
\begin{tabular}{l | c | c | c | c c c}
\toprule
\multirow{2}{*}{Method} & \multirow{2}{*}{Backbone} & 
\multirow{2}{*}{FT} & \multirow{2}{*}{REC} &
\multicolumn{3}{c}{Val set} \\
\cline{5-7}
 & & & & MAE$\downarrow$ & RMSE$\downarrow$ & F1$\uparrow$ \\
\hline
ZSC \cite{Xu2023ZeroShotOC} & ResNet-50 & \cmark & \xmark & 14.84 & 31.30 & - \\
ZSC \cite{Xu2023ZeroShotOC} & Swin-T & \cmark & \xmark & 12.96 & 26.74 & - \\
TFOC \cite{tfoc} & ViT-B & \xmark & \xmark & 16.08 & 31.61 & 0.12 \\
CounTX \cite{countx} & ViT-B-16 & \cmark & \xmark & 11.88 & 27.04 & - \\
CountGD \cite{countgd} & Swin-B & \cmark & \xmark & 9.51 & 22.91 & - \\
GDino \cite{liu2023grounding} & Swin-T & \cmark & \xmark & 9.03 & 21.98 & 0.65 \\
\midrule
GroundingREC \cite{groundingrec} & Swin-T & \cmark & \cmark & 6.80 & 18.13 & 0.68 \\
CAD-GD \cite{cadgd} & Swin-T & \cmark & \cmark & 6.33 & 15.90 & 0.67 \\
\rowcolor{darkgray!10}
\textbf{GroundingDINO + C-REX} & Swin-T & \cmark & \cmark &
\textbf{4.26} & \textbf{12.89} & \textbf{0.73} \\
\bottomrule
\end{tabular}
\caption{Validation set results on REC-8K. FT indicates fine-tuning on REC-8K; REC indicates methods designed specifically for referring expression counting. Best results are in bold.}
\label{tab:rec8k_val}
\end{table}

\begin{table}[!t]
\centering
\begin{tabular}{l | c | c | c | c c c}
\toprule
\multirow{2}{*}{Method} & \multirow{2}{*}{Backbone} &
\multirow{2}{*}{FT} & \multirow{2}{*}{REC} &
\multicolumn{3}{c}{Test set} \\
\cline{5-7}
 & & & & MAE$\downarrow$ & RMSE$\downarrow$ & F1$\uparrow$ \\
\hline
ZSC \cite{Xu2023ZeroShotOC} & ResNet-50 & \cmark & \xmark & 14.93 & 29.72 & - \\
ZSC \cite{Xu2023ZeroShotOC} & Swin-T & \cmark & \xmark & 13.00 & 29.07 & - \\
TFOC \cite{tfoc} & ViT-B & \xmark & \xmark & 17.27 & 32.68 & 0.11 \\
CounTX \cite{countx} & ViT-B-16 & \cmark & \xmark & 11.84 & 25.62 & - \\
CountGD \cite{countgd} & Swin-B & \cmark & \xmark & 11.33 & 30.87 & - \\
GDino \cite{liu2023grounding} & Swin-T & \cmark & \xmark & 8.88 & 21.95 & 0.66 \\
\midrule
GroundingREC \cite{groundingrec} & Swin-T & \cmark & \cmark & 6.50 & 19.79 & 0.69 \\
CAD-GD \cite{cadgd} & Swin-T & \cmark & \cmark & 6.25 & 18.91 & 0.69 \\
\rowcolor{darkgray!10}
\textbf{GroundingDINO + C-REX} & Swin-T & \cmark & \cmark &
\textbf{4.38} & \textbf{17.93} & \textbf{0.75} \\
\bottomrule
\end{tabular}
\caption{Test set results on REC-8K. FT indicates fine-tuning on REC-8K; REC indicates methods designed specifically for referring expression counting. Best results are in bold.}
\label{tab:rec8k_test}
\end{table}

\section{Qualitative Results}
\begin{figure*}[t]
  \centering
  \includegraphics[width=\textwidth]{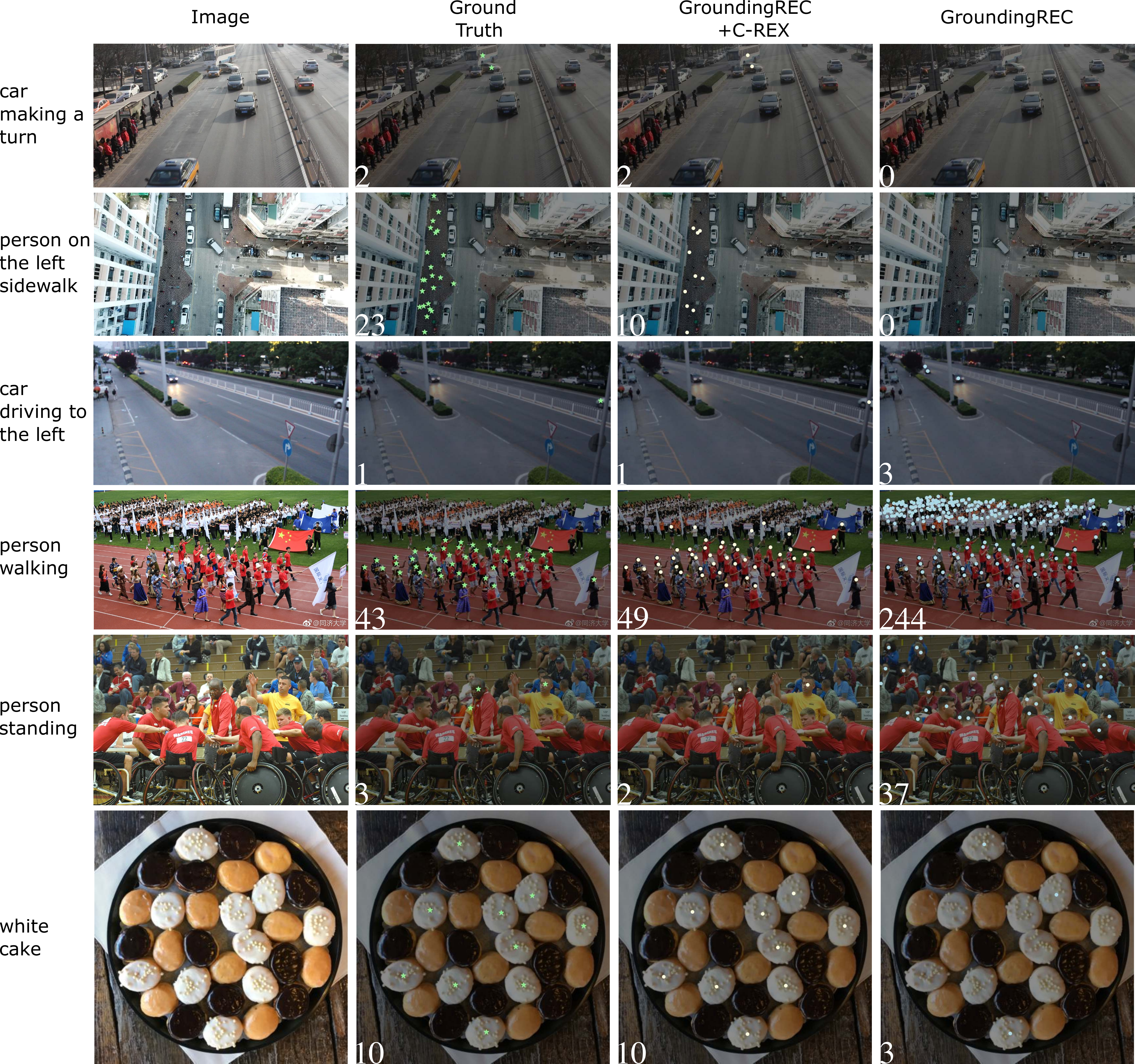}
  \caption{Qualitative comparison of REC predictions for GroundingREC trained with and without C-REX. Across all examples, training with C-REX provides more precise counts and ability to localize a more accurate subset of objects specified by the referring expression. Specifically, with C-REX, GroundingREC can localize instances that it could not handle at all \textbf{(Rows 1, 2)}, and also be more precise and localize precisely the instances that correspond to the RE (``person walking", ``person standing") instead of counting all class instances \textbf{(Rows 3, 4)}.}
  \label{fig:qual_groundingrec}
\end{figure*}

\begin{figure*}[t]
  \centering
  \includegraphics[width=0.9\textwidth]{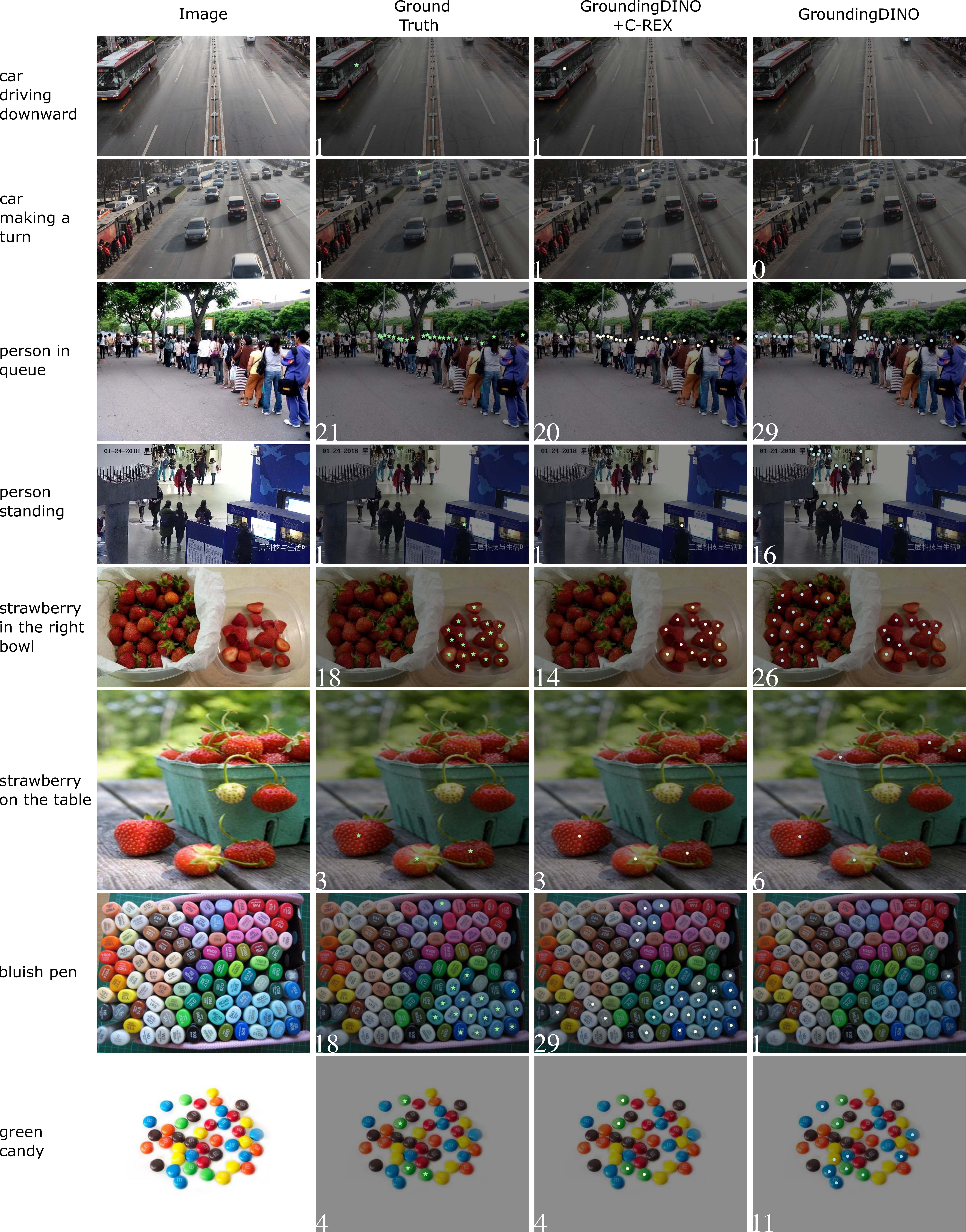}
  \caption{Qualitative comparison of REC predictions for GroundingDINO trained with and without C-REX. Across all examples, training with C-REX provides more precise counts and ability to localize a more accurate subset of objects specified by the referring expression. Specifically, with C-REX, GroundingDINO can localize instances that it could not handle at all \textbf{(Rows 1, 2, 7)}, and also be more precise and localize precisely the instances that correspond to the RE (``strawberry on the table", ``strawberry in the right bowl", ``person in the queue", ``person standing") instead of counting all class instances \textbf{(Rows 3, 4, 5, 6)}.}
  \label{fig:qual_groundingdino}
\end{figure*}

\begin{figure*}[t]
  \centering
  \includegraphics[width=\textwidth]{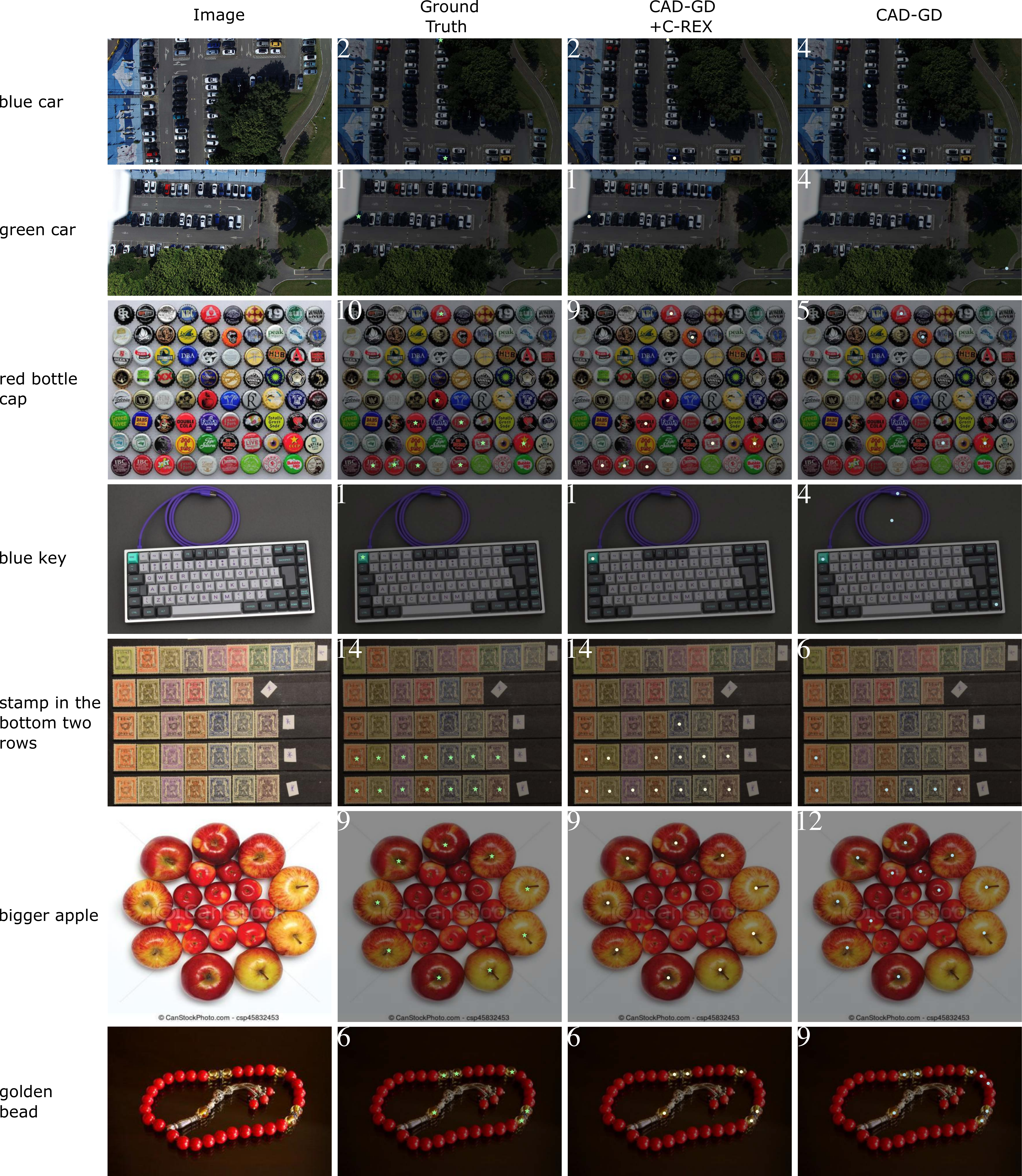}
  \caption{Qualitative comparison of REC predictions for CAD-GD trained with and without C-REX. Across all examples, training with C-REX provides more precise counts and ability to localize a more accurate subset of objects specified by the referring expression. Specifically, with C-REX, CAD-GD can be more precise and localize precisely the instances that correspond to the RE.}
  \label{fig:qual_cadgd}
\end{figure*}

\begin{figure*}[t]
  \centering
  \includegraphics[width=\textwidth]{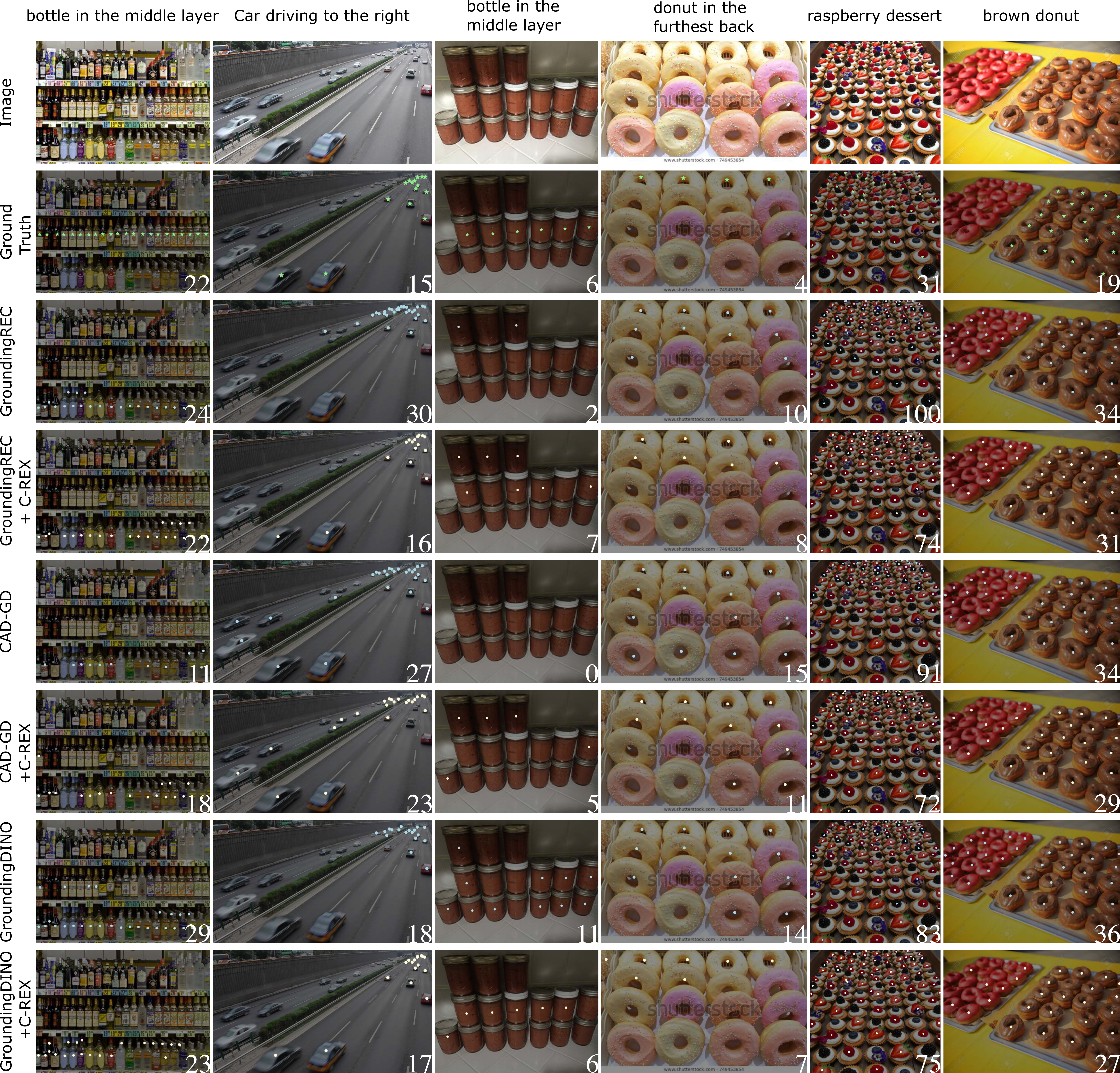}
  \caption{Qualitative comparison of failure cases for REC predictions across different models. The first row shows the input images and the second row visualizes the ground-truth instance annotations for each referring expression. The remaining rows present predictions from each baseline model (GroundingREC, CAD-GD, and GroundingDINO) followed by their corresponding versions with C-REX applied. Each column corresponds to a different referring expression. Across all examples, models trained with C-REX provide more precise counts, even though the instances counted do not correspond to the correct ones. This shows that even when trained with C-REX counting models are still not following the RE properly in some cases.}
  \label{fig:qual_bad}
\end{figure*}

We provide additional qualitative results for all models trained with and without C-REX. Examples for GroundingREC\cite{groundingrec} in Figure \ref{fig:qual_groundingrec}, for GroundingDINO\cite{liu2023grounding} in Figure \ref{fig:qual_groundingdino}, and for CAD-GD\cite{cadgd} in Figure \ref{fig:qual_cadgd}. We also provide examples of failure cases in Figure \ref{fig:qual_bad}.

\section{Comparison with Class-Agnostic Counting models on REC}

Tables \ref{tab:rec8k_val} and \ref{tab:rec8k_test} summarize the performance of prior class-agnostic counting (CAC) models and REC baselines on the REC-8K validation and test sets. Traditional CAC methods, although effective on FSC-147 and other class-agnostic benchmarks, perform substantially worse on REC-8K, reflecting the increased difficulty of counting based on fine-grained linguistic attributes. Integrating C-REX further improves performance, with GroundingDINO + C-REX achieving the strongest results across both validation and test sets. These gains indicate that focusing on visual representation space, rather than relying on text-image alignment, benefits both localization and attribute-aware discrimination.

\label{sec:vs-existing}

\end{document}